# Solving Limited-Memory Influence Diagrams Using Branch-and-Bound Search

**Arindam Khaled** and **Eric A. Hansen**
Dept. of Computer Science and Eng.
Mississippi State University
Mississippi State, MS 39762
{ak697, hansen}@cse.msstate.edu

**Changhe Yuan**
Dept. of Computer and Information Science
Queens College/CUNY
Queens, NY 11367
changhe.yuan@qc.cuny.edu

## Abstract

A limited-memory influence diagram (LIMID) generalizes a traditional influence diagram by relaxing the assumptions of regularity and no-forgetting, allowing a wider range of decision problems to be modeled. Algorithms for solving traditional influence diagrams are not easily generalized to solve LIMIDs, however, and only recently have exact algorithms for solving LIMIDs been developed. In this paper, we introduce an exact algorithm for solving LIMIDs that is based on branch-and-bound search. Our approach is related to the approach of solving an influence diagram by converting it to an equivalent decision tree, with the difference that the LIMID is converted to a much smaller decision graph that can be searched more efficiently.

## 1 Introduction

An influence diagram (ID) is a compact graphical model of a decision problem under uncertainty [5]. In its traditional form, an ID satisfies the assumptions of *regularity* and *no forgetting*, which means that decisions are temporally ordered and each decision is conditioned on all relevant previous observations and decisions. Lauritzen and Nilsson [8] introduced a more general model, called a *limited-memory influence diagram* (LIMID), that allows the regularity and no-forgetting assumptions to be relaxed in order to model a wider range of decision problems. In particular, relaxing the regularity assumption allows modeling of cooperative multi-agent decision problems where one agent is not aware of some or all decisions of another agent. (Note that Howard and Matheson [5] call the regularity assumption the *single decision maker* condition.) Relaxing the no-forgetting assumption allows a decision to be conditioned on a limited number of relevant previous observations and decisions, allowing tradeoffs between the quality of a decision strategy and the complexity of finding it.

Algorithms for solving traditional IDs, such as the join tree algorithm [6], make use of the regularity and no-forgetting assumptions. Thus they cannot be easily generalized to solve LIMIDs. The first algorithm developed to solve LIMIDs, due to Nilsson and Lauritzen [14], is an iterative solution procedure, called single policy updating, that only finds an exact solution under very limited conditions on the structure of the ID; in general, it converges to a locally-optimal solution. The first exact and general algorithm for solving LIMIDs, developed by de Campos and Ji [1], reformulates a LIMID as a credal network inference problem that can be solved by mixed integer programming. Maua and de Campos [11, 10] recently developed a more efficient exact algorithm for solving LIMIDs based on variable elimination, called multiple policy updating.

In this paper, we introduce another exact algorithm for solving LIMIDs. Our approach builds on the work of Yuan et al. [24], who describe a branch-and-bound search algorithm for solving a traditional ID and show that it can outperform other approaches to solving IDs for multistage decision problems. We adopt the same branch-and-bound approach, but with some important differences. The branch-and-bound algorithm for solving a traditional ID is a tree-search algorithm in which each path through the decision tree represents perfect memory of a particular history of decisions and observations, in keeping with the no-forgetting assumption of a traditional ID. By contrast, our branch-and-bound algorithm for solving LIMIDs searches in a much smaller search *graph* in which different paths to the same node of the graph represent different histories where the differences are not "remembered." By collapsing the search tree into a smaller search graph in which fewer histories are distinguished, the branch-and-bound approach can solve the optimization problem for LIMIDs much more efficiently. That is, the new graph-search techniques we introduce leverage the opportunities for faster strategy computation provided by the LIMID model. We also develop new techniques for probabilistic inference and bounds computation in the search graph that further enhance this approach to solving LIMIDs. Experimental results demonstrate the effectiveness of this approach.

## 2 Background

We begin with a review of limited-memory influence diagrams and previous work on solving influence diagrams using branch-and-bound search.

### 2.1 Influence diagrams

An influence diagram (ID) represents a decision problem by a directed acyclic graph with three types of nodes: chance nodes, decision nodes, and utility nodes. The chance nodes represent random variables, $\mathbf{X} = \{X_1, \ldots, X_n\}$, where each random variable $X_i \in \mathbf{X}$ has an associated domain of possible values, $dom(X_i)$. The decision nodes represent decision variables, $\mathbf{D} = \{D_1, \ldots, D_m\}$, where each decision variable $D_i \in \mathbf{D}$ has an associated domain of possible values, $dom(D_i)$, called actions. For both random variables and decision variables, all domains are assumed to be non-empty and finite. The utility nodes represent (local) utility functions, $\mathbf{U} = \{U_1, \ldots, U_l\}$, that express a decision maker's preferences. By convention, an ID shows chance nodes as circles, decision nodes as squares, and utility nodes as diamonds.

The edges of the graph characterize dependencies among nodes and have a different meaning depending on their destination. Incoming edges to a chance node indicate probabilistic dependence. As in a Bayesian network, each random variable $X_i \in \mathbf{X}$ has an associated conditional probability table $P(X_i|\pi(X_i))$, where $\pi(X_i)$ denotes the set of parent variables of $X_i$ and $dom(\pi(X_i))$ denotes the set of possible instantiations (or states) of the parent variables. Incoming edges into decision nodes are informational, and the parent variables $\pi(D_i)$ of a decision variable $D_i \in \mathbf{D}$, called the *information variables* of the decision, are the variables whose values are known to the decision maker at the time the decision is made. An instantiation of the information variables is called an *information state*, and $dom(\pi(D_i))$ denotes the set of all information states for the decision variable $D_i$. Incoming edges into utility nodes indicate functional dependence, and a utility function $U_i : \Omega_{\pi(U_i)} \to \Re$ maps each state of the parent variables $\pi(U_i)$ to a utility value that represents the preference of the decision maker. It is assumed that utility nodes do not have children and the joint utility function $\mathcal{U}$ is additively decomposable such that $\mathcal{U} = \sum_{U_i \in \mathbf{U}} U_i$.

An ID is solved by finding a strategy that maximizes expected utility. A *strategy* $s = \{\delta_{D_i}|D_i \in \mathbf{D}\}$ is a set of policies, one for each decision variable, where a *policy* $\delta_{D_i} : dom(\pi(D_i) \to dom(D_i)$ is a mapping from the information states of a decision variable to the possible actions for that decision variable. A strategy $s$ induces a joint probability distribution $P_s$ over $\mathbf{X} \cup \mathbf{D}$, as follows,

$$P_s(\mathbf{X} \cup \mathbf{D}) = \Pi_{X_i \in \mathbf{X}} P(X_i|\pi(X_i)) \cdot \Pi_{D_j \in \mathbf{D}} P_s(D_i|\pi(D_i)), \quad (1)$$

where

$$P_s(d|\pi(D_i)) = \begin{cases} 1 & \text{if } \delta_{D_i}(\pi(D_i)) = d, \\ 0 & \text{otherwise.} \end{cases} \quad (2)$$

The expected utility of a *strategy* $s$ is defined as

$$EU(s) = \sum_{c \in (\mathbf{X} \cup \mathbf{D})} P_s(c) \sum_{U_i \in \mathbf{U}} U_i(\pi(U_i)) \quad (3)$$

$$= \sum_{U_i \in \mathbf{U}} \sum_{\pi(U_i)} P_s(\pi(U_i)) \cdot U_i(\pi(U_i)), \quad (4)$$

where $c \in (\mathbf{X} \cup \mathbf{D})$ denotes a particular configuration (or instantiation) of the variables of the ID. A strategy $s^*$ is optimal if $EU(s^*) \geq EU(s)$ for all strategies $s$.

For an ID that satisfies the regularity assumption, the decision variables are temporally ordered. Suppose there are $n$ decision variables $D_1, D_2, \ldots, D_n$. The decision variables partition the random variables in $\mathbf{X}$ into a collection of disjoint sets $\mathbf{I_0}, \mathbf{I_1}, \ldots, \mathbf{I_n}$. For each $k$, where $0 < k < n$, $\mathbf{I_k}$ is the set of random variables that must be observed between $D_k$ and $D_{k+1}$. $\mathbf{I_0}$ is the set of initial evidence variables that must be observed before $D_1$. $\mathbf{I_n}$ is the set of variables left unobserved when decision $D_n$ is made. Therefore, a partial order $\prec$ is defined on the ID over $\mathbf{X} \cup \mathbf{D}$, as follows:

$$\mathbf{I_0} \prec D_1 \prec \mathbf{I_1} \prec \ldots \prec D_n \prec \mathbf{I_n}. \quad (5)$$

When the *no-forgetting* assumption is satisfied, all information variables of earlier decisions are also information variables of later decisions. We call these past information variables the *history*, and, for convenience, we assume that there are *explicit* edges (called information arcs) from history information variables to decision variables. As the number of variables in the history grows, however, the domain of the policy for each decision variable increases exponentially. Methods for structural analysis of relevance have been developed that can distinguish *requisite* observations from those that are irrelevant, and remove information arcs that are not necessary for computation of the optimal strategy [21, 13, 8]. This preprocessing step can be performed prior to any numerical evaluation of the LIMID.

Even if information arcs from irrelevant variables are removed and only relevant information variables are considered for each decision, the domain of the policy for a decision variable may grow exponentially in the number of relevant information variables in the history. Limited-memory influence diagrams [8] address this problem by allowing information arcs from relevant variables to be removed. When an information variable for an earlier decision is not an information variable for a later decision, it means the no-forgetting assumption is violated. If there is not an information arc from an earlier decision variable to a later decision variable, it means the regularity assumption is violated. We use the following example to illustrate the properties of a LIMID.

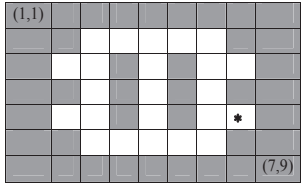

Figure 1: A maze with goal state marked by starred cell.

## 2.2 Example

Consider a maze navigation problem that can be modeled as an ID [4, 14, 24]. In the maze shown in Figure 1, white cells represent spaces where navigation is possible, and shaded cells represent walls. A robot is initially placed in one of the white cells and its objective is to reach the goal state marked by a star. At each stage, the robot can choose to move to any one of its neighboring cells, or it can stay in its current location. The effect of the robot's attempts to move are stochastic. It moves to the intended neighboring cell with a probability of 0.89 and fails to move with probability 0.089. The probability of sideways movement is 0.01 in each of two possible directions, and the probability of backward movement is 0.001. A move towards a wall has a probability of zero to succeed, and the remaining non-zero probabilities are normalized. The robot's sensors provide incomplete information about the robot's location. They detect neighboring walls, but since more than one location can share the same configuration of neighboring walls, observations do not unambiguously identify the current location. The expected utility received by the robot corresponds to the probability of successfully reaching the goal state by the final stage of the problem. If the robot is in the goal state at the final stage, it receives a utility value of 1; otherwise, it receives a value of 0.

The ID shown in Figure 2(a) represents a two-stage version of the maze navigation problem. The random variables $x_i$ and $y_i$ represent the coordinates of the location of the robot at stage $i$. The random variables $\{ns_i, es_i, ss_i, ws_i\}$ are the sensor readings in four directions at stage $i$. The decision variable $d_i$ represents one of the possible actions taken by the robot at that stage. The ID shown in Figure 2(a) is a traditional ID that satisfies the regularity and no-forgetting assumptions. Figure 2(b) shows a LIMID for which the no-forgetting assumption is not satisfied. In this case, a decision is conditioned on all past decisions as well as the present states of the information variables when the decision is made, but not on the information variables for any previous stages. In other words, the robot makes decisions based on its current sensor readings only, without considering any previous sensor readings. Although the IDs shown in Figure 2 represent two-stage decision problems, they are easily extended for any finite number of stages.

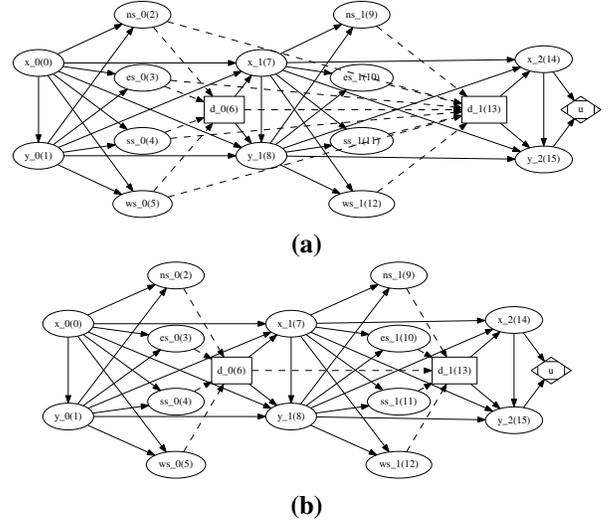

Figure 2: (a) An influence diagram with no-forgetting and (b) a LIMID, both for the maze navigation problem.

## 2.3 Branch-and-bound solution method

Several exact methods have been developed to solve IDs that satisfy the regularity and no-forgetting assumptions. IDs were first solved by converting them to a decision tree [5]. Subsequently, methods were developed that solve an ID directly [19], or by converting it to some other graphical form, such as a junction tree [20, 6]. For an ID that has been converted to a decision tree, the traditional solution method is the "average-out and fold-back" algorithm [17]. However, improved performance can be achieved using a search algorithm that traverses the decision tree beginning from the root and prunes branches with zero probability [16, 9]. The performance of this approach can be improved further by using bounds to prune the tree, and the results of this paper build on our earlier work on solving IDs using depth-first branch-and-bound search [24]. In this approach, the decision tree corresponds to an AND/OR search tree in which AND nodes correspond to information variables (i.e., chance nodes that have informational arcs into a decision node), OR nodes correspond to decision nodes, and leaf nodes correspond to utility nodes. A path from the root of the search tree to a leaf node corresponds to an instantiation of the information and decision variables of the ID. When traversed by a depth-first search algorithm, the tree is generated "on the fly" and only part of the AND/OR tree needs to be in memory at any one time.

Two issues must be addressed to develop an effective branch-and-bound algorithm. We need bounds to prune the search tree and we need an efficient method for computing posterior probabilities. As we discuss next, our approach to both issues involves construction of a secondary ID we call a *relaxed influence diagram*.

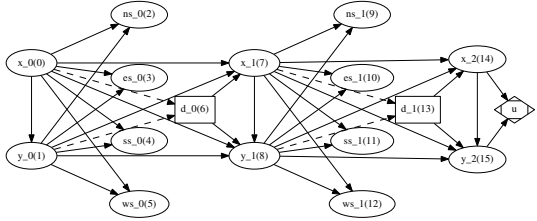

Figure 3: Relaxed ID for two-stage maze problem.

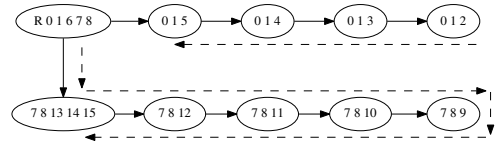

Figure 4: Strong join tree for the relaxed ID in Figure 3.

**Bounds computation** We need bounds on the values of OR nodes in order to prune the AND/OR search tree. Assuming that we are maximizing expected utility, the best value computed so far for any branch of an OR node (i..e, for any action) serves as a lower bound on the optimal value of the OR node. To compute upper bounds, we adopt an approach proposed by Nilsson and Höhle [14]. In this approach, upper bounds are computed by solving a secondary ID that is usually much easier to solve, which we call a relaxed ID. The relaxed ID is created by adding information variables to the original ID that provide the decision maker with additional information (ensuring that the solution of the relaxed ID is an upper bound on the solution of the original ID), while allowing the ID to be simplified by removing *non-requisite* arcs (ensuring that it is much easier to solve than the original ID). Recall that an information arc is *non-requisite* [8, 13] for a decision node $D$ if

$$I_i \perp (\mathbf{U} \cap d(D))|D \cup (\pi(D) \setminus \{I_i\}), \qquad (6)$$

where $d(D)$ and $\pi(D)$ are the descendants and parents of $D$, respectively, and $\perp$ denotes conditional independence. A *reduction* of an ID is obtained by deleting all *non-requisite* information arcs [14]. Ideally, we want to add information variables that make some or all of the information arcs for each decision node *non-requisite*, and also make the ID easier to solve. The optimal policy for a decision variable, $D_j$, depends on a set of information variables, $N_j$, or else it is history-independent. This set $N_j$ can be described as the current state of the decision problem, such that if the decision maker is informed of this state, it does not need to know of any past history to find an optimal policy; in this respect, it fulfills the Markov property and can be said to provide perfect information. $N_j$ is called a *sufficient information set* (SIS) of $D_j$ [24].

Thus the relaxed ID is created in two steps. First, the SIS for each decision is computed in reverse time order, $D_n, ..., D_1$, making each SIS the information variables for its corresponding decision variable. Second, *non-requisite* arcs are removed from the ID. Consider the LIMID in Figure 2(b) as an example. The SIS for $d_1$ is found as $\{x_1, y_1\}$. The SIS for $d_0$ is computed to be $\{x_0, y_0\}$. By making $\{x_1, y_1\}$ and $\{x_0, y_0\}$ information variables for $d_1$ and $d_0$, respectively, and after removing the non-requisite arcs, a relaxed ID is obtained which is shown in Figure 3.

**Incremental join tree evaluation** The join tree algorithm, an efficient method for probabilistic inference in a Bayesian network, also provides an efficient method for solving an ID [6]. For optimization problems that can be solved by depth-first branch-and-bound search, Yuan and Hansen [23] describe an incremental version of the join tree algorithm. First developed for a branch-and-bound algorithm for solving the MAP problem, it is used in the branch-and-bound algorithm for solving traditional IDs [24], and we will use it to extend the branch-and-bound approach to LIMIDs. It assumes a static ordering of the variables to be instantiated, and leverages the observation that when only one new variable is instantiated at a time during forward traversal of a branch of a search tree, it is only necessary to perform message passing once along this path in the join tree, broken into separate steps for each instantiating variable. To allow efficient backtracking, the clique and separator potentials of the join tree that are changed during forward traversal are cached in the order that they are changed. During backtracking, the cached potentials can be used to efficiently restore the join tree to its previous state.

A strong join tree constructed from the relaxed ID is used to compute both probabilities and upper bounds for the AND and OR nodes in the search graph. Figure 4 shows a strong join tree for the relaxed ID of Figure 3. We select an order of variable elimination from the join tree that is an extension of the elimination ordering of the ID. Note that one of the partial orders for the LIMID shown in Figure 2(b) is

$$\{ns_0, es_0, ss_0, ws_0\} \prec \{d_0\} \prec \{ns_1, es_1, ss_1, ws_1\}$$
$$\prec \{d_1\} \prec \{x_0, y_0, x_1, y_1, x_2, y_2, u\}, \qquad (7)$$

Any order of variable expansion for the join tree that satisfies the constraints in Equation 7 can be used for *incremental join tree evaluation*. The order suggests which variable to expand/instantiate next. For example, after $ns_0$ is expanded, the only message that needs to be sent to obtain $P(es_0|ns_0)$ is the message from clique $(0, 1, 2)$ to $(0, 1, 3)$. If the following order for the maze problem is selected

$$ns_0, es_0, ss_0, ws_0, d_0, ns_1, es_1, ss_1, ws_1, d_1, x_0, y_0,$$
$$x_1, y_1, x_2, y_2 \qquad (8)$$

then an incremental message-passing scheme can be used in the direction of the dashed arc in Figure 4 in one downward pass of the depth-first search to compute probabilities and upper bounds.

## 3 AND/OR search graph

We next describe how to formulate the problem of solving a LIMID as an AND/OR graph search problem. There are two main differences between the depth-first-branch-and-bound (DFBnB) algorithm for solving LIMIDs that we develop in the rest of the paper, and the DFBnB algorithm for solving traditional IDs. The first difference is that a LIMID is solved by searching in an AND/OR graph instead of an AND/OR tree. Since a decision maker with limited memory is not able to distinguish all histories, the search space for solving a LIMID is a graph in which different paths that represent different histories can lead to the same OR node because the differences between the histories are not remembered. A second difference is that the message-passing scheme used by the incremental join tree algorithm to compute bounds and probabilities requires some revisions for search in an AND/OR graph instead of an AND/OR tree. We discuss the first difference in Section 3.1 and the second in Section 3.2.

First we review some basic concepts about the AND/OR search space for the decision problem represented by an ID. Recall that AND nodes represent random variables and OR nodes represent decision variables. Any arc emitting from an AND node has a probability attached to it; the sum of all the probabilities associated with the arcs of an AND node is 1.0. Each arc emitting from an OR node represents a decision alternative. The leaf nodes of the search graph have utility values attached to them that are derived from the utility nodes of the ID. The *valuation function* for each node is defined as follows: (a) for a leaf node, the value is its utility value, (b) for an AND node, the value is computed by multiplying the probability associated with each outgoing arc by the utility value of the corresponding child node and then summing these values, and (c) for an OR node, the value is the maximum of the utility values of the child nodes. We use this valuation function to determine the optimal strategy for an ID. We represent a strategy for an ID as a *strategy graph*, which is a subgraph of an AND/OR graph that is defined as follows: (a) it consists of the root of the AND/OR graph; (b) if a non-terminal AND node is in the strategy graph, all its children are in the strategy graph; and (c) if a non-terminal OR node is in the strategy graph, exactly one of its children is in the strategy graph. Given an AND/OR graph that represents all possible histories and strategies for an ID, the decision problem is solved by finding a strategy graph with the maximum value at the root, where the value of the strategy graph is computed based on the valuation function.

The optimal strategy for an ID can always be found by searching in an AND/OR tree. But the tree representation of the problem is inefficient if it contains many repeated subtrees that represent the same subproblem. For a LIMID, the number of repeated subtrees will be much greater than for a traditional ID because many different histories will lead to the same subproblem in which a decision must be made based on limited information that represents only part of the history. In the following, we describe how to convert an AND/OR tree representation of the decision problem for a LIMID into an equivalent AND/OR graph by merging OR nodes that are generated from different histories but represent the same subproblem.

In our approach, we (slightly) limit the class of LIMIDs we consider by assuming the following: if a decision maker is aware of the value of an information variable and then "forgets" it, it cannot recall the forgotten information later on. We call this assumption the *no-recalling-forgotten-information* rule. It is difficult to imagine any realistic decision problem that violates this assumption. But we make the assumption explicit because the LIMID model does not rule out such cases. The assumption simplifies our approach to recognizing and merging equivalent OR nodes.

We also modify the usual definition of an AND/OR graph by introducing a special kind of AND node that allows multiple decisions to be considered in parallel if they represent scenarios that are conditionally independent given that some variables have already been observed. We call this new kind of node a *special AND* (SAND) node. We discuss SAND nodes further in Section 3.1.3.

### 3.1 Context-based merging of OR nodes

To construct an AND/OR graph instead of an AND/OR tree for solving LIMIDs, the main idea is to merge multiple OR nodes (i.e., decision nodes) that represent the same decision scenario into a single OR node. Our approach uses the concept of the context of an OR node. The *context* of an OR node is defined as the joint state of the information variables remembered by the current decision variable along with the states of previously observed decisions that will influence the descendant utilities of this decision. More formally, the context of an OR node is a set $C_{D_i} = S_{I_{i-1}} \cup S_{D_{i-1}} \cup S_{D_{u(i-1)}}$, where $S_{I_{i-1}}$, $S_{D_{i-1}}$, and $S_{D_{u(i-1)}}$, respectively are the sets of states of the random and decision variables remembered by $D_i$, and the set of states of the previously expanded decision variables that will influence the descendent utilities of $D_i$.

For example, Figure 5 shows a partial AND/OR graph for the LIMID in Figure 2. Note that the AND/OR graph is condensed for ease of illustration as the individual random variables for the sensors are grouped together into one AND layer. In our actual AND/OR graph, they correspond to multiple AND layers. The context of the bottom-right OR node in the Figure is $\{1, 1, 1, 1, 0\}$, where $\{1, 1, 1, 1\}$ is the set of present information states, and $\{0\}$ is the last action taken. The previous sensor readings are totally forgotten. That is why the two paths starting from the root converge to this OR node.

In our context-based approach to merging OR nodes, we distinguish three cases; sequential decisions, cooperative decisions, and conditionally-independent decisions. We explain the differences below.

### 3.1.1 Sequential decisions

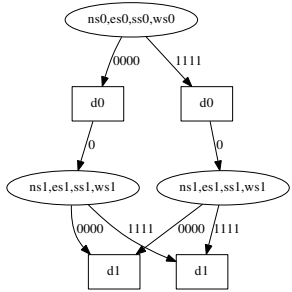

Figure 5: Partial AND/OR graph with merged OR nodes.

A pair of decisions is *sequential* if there is an obvious temporal ordering between them. If all decision pairs in a LIMID are sequential then the elimination order of Equation 5 applies. Figure 2(b) shows an example of a LIMID where all decision pairs are sequential. Duplicate OR nodes are detected using the definition of context provided above and illustrated by the example in Figure 5. Contexts are rather straightforward for sequential decisions given the no-recalling-forgotten-information assumption. Without this assumption, much more complex book-keeping would be needed to keep track of contexts during the search.

Note that in Figure 5, the AND nodes contain multiple random variables to simplify the picture. In the AND/OR search graph created by our implementation, each random variable is represented by a separate AND node, and they are expanded one at a time by the search algorithm.

### 3.1.2 Cooperative decisions

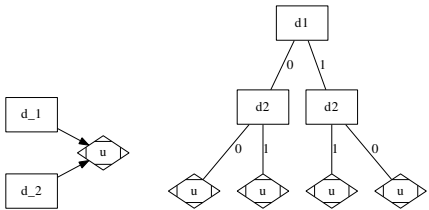

Figure 6: A LIMID with a cooperative-decision pair (left), and its corresponding AND/OR graph (right).

We next consider the case where multiple, simultaneous decisions need to be considered to evaluate a value function. We call this case a *cooperative decision* because the decision makers need to cooperate to select the combination of actions that results in an optimal utility for every possible decision scenario. Unlike the sequential decision scenario, there exists no obvious elimination ordering between decision pairs in this case. Figure 6 shows an example of a cooperative decision pair. In this example, if $d_2$ is realized after $d_1$ in the elimination order during the generation of the AND/OR graph, then $d_1$ needs to be included in $d_2$'s context. The AND/OR graph for this LIMID is also shown in Figure 6. Note that all the OR nodes in this graph are unique.

### 3.1.3 Conditionally-independent decisions

It is possible that multiple decisions can be expanded in parallel given that some variables have been realized. An example is shown in Figure 7. In this example, the sets of decisions $\{D_1\}$, and $\{D_2\}$ can be expanded in parallel given that the set of variables(s), $\{a\}$, has been observed. We call this situation a *conditionally-independent decision scenario* (CIDS). A CIDS can be determined from the LIMID itself or the strong join tree of the LIMID [7, 21].

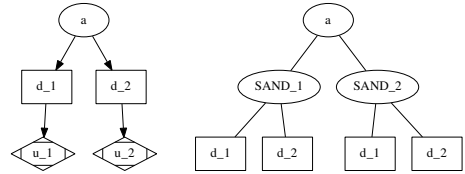

Figure 7: A LIMID with CIDS(left), and its corresponding (partial) AND/OR graph (right).

To solve a LIMID that includes a CIDS, we introduce a new type of node in the AND/OR graph, called a *special AND* or SAND node. When it is realized that multiple decisions can be made in parallel after some nodes in the search graph have been expanded, a SAND node is introduced (in the search graph). A SAND node is different from a regular AND node in two ways: the total number of children for a SAND node is the number of sets of decisions that can be expanded in parallel, and the weight attached to each arc is 1.0. Unlike the regular AND/OR graph search where the rest of the unexpanded nodes are considered for expansion once a node (AND or OR) is expanded for each branch of the AND node, each branch of the SAND node will expand a subset of unexplored nodes for a LIMID with CIDS(s). These subsets are determined by the elimination order of decisions presented by the join tree: each subset will contain the set of decisions and their respective information variables that need to be expanded in sequence in the future. For example, for the LIMID shown in Figure 7, once variable $a$ is expanded, a SAND node, denoted $SAND_1$, will be expanded. This node will have two branches – in one branch the set of decision(s), $\{D_1\}$, will be expanded; the set $\{D_2\}$ will be expanded in the other branch, as illustrated in Figure 7.

### 3.1.4 Implementation

So far, we have given rules for merging OR nodes but have not discussed their implementation. For each OR node in the graph, we store both its context and utility value (once it is calculated). When an OR node is ready to be generated, its context is calculated and then checked against the contexts of existing OR nodes (which we store in a hash table, described below). If a previously-generated OR node has the same context, the new OR node is not generated. Instead, the existing OR node receives an additional arc from the parent of the new OR node.

For duplicate checking, all generated OR nodes are stored in a hash table indexed by their context. (In cases in which we can decompose the decision problem into well-defined stages, there can be a separate hash table for each stage of the problem.) In our implementation, the context is represented by a string that contains the states of the variables of that decision's context, concatenated by commas. As the search progresses, the context string for each decision node can grow and potentially slow the duplication detection process.

### 3.1.5 Discussion

The concept of solving an ID by searching in an AND/OR graph, instead of an AND/OR tree, is not new. In the literature on IDs, the process of converting a decision tree to an equivalent graph in which identical subtrees are merged is referred to as *coalescence* [15, 22]. Automating coalescence in the decision tree framework is considered difficult and computationally expensive, however, and solutions are sometimes hand-crafted. A context-based approach to merging OR nodes has been proposed before, for probabilistic inference in Bayesian networks [2] and solving an ID [9]. The primary difference between our approach and previous work is that our approach applies to LIMIDs.

Note that our approach to context-based merging does not necessarily produce the most concise AND/OR graph. A more concise graph could be found by directly comparing probability distributions to detect duplicate OR nodes. We define $R_{D_i}$ as the set of variables that is considered for expansion once the information variables, $I_{i-1}$, for a decision variable, $D_i$, are expanded. Formally, $R_{D_i} = D_i \cup I_i \cup D_{i+1} \cup ... \cup I_{n-1} \cup D_n$. Once the information variable set, $I_{i-1}$, is expanded, the probability distribution of $R_{D_i}$ given $D_i$'s context, $\mathbb{P}(R_{D_i}|C_{D_i})$, could be used to detect duplicate decision scenarios. If multiple decision scenarios share the same distribution then these OR nodes can be collapsed into a single OR node, since they share a subgraph [22]. Although comparing probability distributions in order to merge nodes could generate a more compact AND/OR search graph, it is computationally expensive, and an approach that relies on context-based merging appears to be more practical.

## 3.2 Incremental probabilities and bounds

We next consider how to modify the incremental join tree algorithm to compute the probabilities and bounds needed by the AND/OR graph search algorithm for solving LIMIDs. Consider the LIMID shown in Figure 2.2(b) as an example, and the relaxed LIMID of Figure 4. We can use the join tree of the relaxed LIMID to compute both the probabilities and bounds needed for the AND/OR graph search. (We can use it to compute probabilities for the AND nodes of the search graph because the same set of actions transforms both the original LIMID and the relaxed LIMID into the same Bayesian network. Adding information arcs to create a relaxed ID only changes the expected utility of the network.) Note that the join tree does not have a clique that contains all four variables; they are in different cliques. Thus we consider the expansion of these variables one by one, generating an AND/OR graph with four layers of AND nodes followed by a layer of OR nodes, as shown in Figure 5.

The AND/OR search graph is generated on-the-fly during the branch-and-bound search. To make this process efficient, we follow the elimination order given by Equation 8 to generate the nodes of the search graph and calculate probabilities. The probabilities for the AND nodes corresponding to variables $\{ns_0, es_0, ss_0, ws_0\}$ can be calculated by sending messages in the following order of cliques: $(0, 1, 2)$, $(0, 1, 3)$, $(0, 1, 4)$, and $(0, 1, 5)$. Given limited memory, however, calculation of probabilities for the information variables of the next decision needs to be modified.

When a decision stage is considered for expansion, we consider whether the decision variable, $D_i$, recalls anything from the past. If it does, then the cliques hosting these recalled variables along with the clique, $clq_0$, that hosts the first information variable for $D_i$ are identified. Then we devise a message-passing scheme that sets evidence for the cliques of the recalled variables and passes messages towards the clique, $clq_0$. To perform these message propagations, we use a set of temporary potentials, one assigned for each clique and separator, which are initialized to the clique potentials obtained from the initial collection and distribution process of the join tree at the beginning of this process. For our example, when the first information variable, $ns_1$, for $d_1$ is about to be expanded, we set evidence to the clique $(0, 1, 6, 7, 8)$ with the current state of $d_0$, and pass in the direction of the clique $(7, 8, 9)$. Once the clique $(7, 8, 9)$ receives this message, it sets its current potential to this newly obtained potential. The rest of the information variable expansion process follows the incremental join tree evaluation method proposed in [24]. When backtracking from a decision, we backtrack to the clique hosting the last information variable expanded for the previous expanded decision. The space requirement for our join tree evaluation approach is $O(N)$ if $N$ is the space required for the evaluation approach proposed in [24].

After the AND nodes of $\{ns_0, es_0, ss_0, ws_0\}$ are generated for this example, we need to generate the OR node $d_0$ and corresponding upper bounds. The subset of information variables for a decision $d_i$ is used to compute the upper bound for $d_i$. In our example, we do not need to set the states of $\{ns_0, es_0, ss_0, ws_0\}$ as evidence during the computation of expected utility values for $d_0$ because the states of the information variables for $d_0$ are not remembered for future decisions. In summary, the basic idea for handling information forgetting in a LIMID is to only send a message to a future decision for calculating probabilities and utility values if the message contains information variables that are remembered by the future decision.

### 3.3 Optimality of the algorithm

It is not difficult to prove that our AND/OR graph search algorithm finds an optimal strategy. If we do not merge OR nodes, we have an AND/OR tree, and so we first show that the strategy found by an AND/OR tree search is optimal.

**Lemma 1** *A DFBnB AND/OR tree search algorithm finds an optimal strategy for a LIMID.*

*Proof:* Since all the possible policies for each decision node are examined by the search, it must converge to an optimal strategy once the search ends. □

We then argue that merging OR nodes preserves optimality.

**Theorem 1** *A DFBnB AND/OR graph search algorithm finds an optimal strategy for a LIMID.*

*Proof:* Since the subgraphs below any OR nodes that represent the same decision scenario are identical and have the same utility, merging the OR nodes preserves optimality. □

## 4 Experimental evaluation

We tested the performance of our algorithm in solving the maze problem described in Section 2.2, as well as a classic finite-horizon DEC-POMDP, and several randomly-generated LIMIDs. Experiments were performed on a Windows PC with a Pentium i3 processor and 3GB of RAM.

Tables 1, 2 and 3 show all results in the same format. The column labeled "$(d, c, u)$" gives the number of decision nodes, chance nodes, and utility nodes, respectively, in the LIMID, and the column labeled "SG" gives the size of the optimal strategy graph measured as the total number of AND and OR nodes it contains. The remaining columns measure the efficiency of the search algorithm. The column labeled "Pruned" gives the number of times a branch of the AND/OR graph was pruned in solving the LIMID, the column labeled "Merged" gives the number of times two OR nodes were merged, and the column labeled "Time" gives the time needed to solve the problem.

| $(d, c, u)$ | SG | Pruned | Merged | Time |
|---|---|---|---|---|
| $(2, 14, 1)$ | 495 | 124 | 528 | $109ms$ |
| $(3, 20, 1)$ | 951 | 364 | 2112 | $421ms$ |
| $(4, 26, 1)$ | 1407 | 688 | 4224 | $952ms$ |
| $(5, 32, 1)$ | 1,863 | 2,848 | 18,480 | $4s$ |
| $(6, 38, 1)$ | 2,319 | 9,412 | 61,776 | $17s$ |
| $(7, 44, 1)$ | 2,775 | 25,768 | 168,960 | $53s$ |
| $(8, 50, 1)$ | 3,231 | 90,400 | 593,472 | $3m30s$ |
| $(9, 56, 1)$ | 3,687 | 309,184 | 2,027,520 | $13m21s$ |
| $(10, 62, 1)$ | 4,143 | 1,058,908 | 6,939,504 | $50m17s$ |

Table 1: Results for maze navigation LIMID.

### 4.1 Maze navigation

Table 1 shows results for the maze navigation problem of Section 2.2 when the number of stages is varied from two through ten. The LIMIDs for this problem satisfy the regularity assumption (there is one decision node per stage), but not the no-forgetting assumption (observations are not remembered after the current stage). Of the total number of branches pruned, approximately 40% are pruned based on bounds; the remaining 60% are pruned because the probability of the branch is zero.

### 4.2 Multi-agent tiger behind door

Table 2 shows results for a finite-horizon DEC-POMDP that represents cooperative multiagent decision making under uncertainty [12]. In this problem, there are two doors, one on the left and one on the right, and two agents. Behind one door is a tiger and behind the other is treasure. Each agent has a choice of three actions: it can open a door on the left or right, or it can listen for the tiger. If an agent hears the tiger behind one of the doors, the tiger is actually there with probability 0.85. At each stage, the agents must each choose an action without knowing what the other agent will choose; thus the regularity assumption is not satisfied. Each agent remembers its previous actions and observations, but is unaware of the other agent's observations. The agents receive better rewards if they coordinate their actions: the reward for opening the door with treasure is greater is both agents open the door together, and the penalty for opening the door with the tiger is less severe if they open that door together. There is a small cost for the listen action.

| $(d,c,u)$ | SG | Pruned | Merged | Time |
|---|---|---|---|---|
| $(4, 6, 2)$ | 15 | 12 | 6 | $15ms$ |
| $(6, 9, 3)$ | 39 | 35 | 24 | $62ms$ |
| $(8, 12, 4)$ | 87 | 737 | 576 | $530ms$ |
| $(10, 15, 5)$ | 351 | 9,529 | 6,984 | $7s$ |
| $(12, 18, 6)$ | 1,599 | 42,559 | 31,602 | $49s$ |
| $(14, 21, 7)$ | 4,047 | 288,516 | 214,170 | $5m31s$ |
| $(16, 24, 8)$ | 10,023 | 2,328,571 | 1,763,904 | $53m21s$ |

Table 2: Results for multi-agent tiger LIMID.

Table 2 shows results for this problem when the number of stages is varied from two through eight. Our algorithm solves the problem optimally for eight stages in less than an hour. The provably optimal solution reported in [18] is for up to four stages, found by dynamic programming [3]. For this search problem, there are no zero-probability branches. All branches are pruned based on bounds.

### 4.3 Randomly-generated LIMIDs

We also tested our algorithm on a set of randomly-generated LIMIDs. The LIMIDs were created to have between 10 and 20 stages, with one decision node, one utility node, and between 3 and 6 chance nodes per stage. For each stage, half (and at least 2) of the nodes are selected to be information variables of the decision variable. The utility function for each stage is a function of the decision variable and two randomly-selected random variables from that stage, and potentially the decision variable from the previous stage. Once nodes are generated for all stages, we generate additional informational arcs as follows. For the decision variable of each stage $k$ beginning from the second stage and continuing until the last stage, we add informational arcs from half (and at least 2) of the information variables of the previous stage, selected randomly. This method of adding informational arcs ensures that the no-recalling-forgotten-information rule is satisfied. When adding arcs, we make sure that chance nodes with no children become the information variables for the decision node first so that there are no barren nodes. Each random and decision variable has from 2 to 4 states, and the probabilities for the random nodes are assigned from an uniform probability distribution. The utility values range from $-20$ to $20$.

Table 3 only shows results for a selection of LIMIDs for which the treewidth of the relaxed LIMID does not exceed 12. The LIMIDs solved are larger than those for the maze and tiger problems. For these randomly-generated LIMIDs, not all previous actions are remembered, which appears to make both duplicate detection and probabilistic inference using the incremental join tree algorithm faster.

| (d,c,u) | SG | Pruned | Merged | Time |
|---|---:|---:|---:|---:|
| (10, 46, 10) | 33,463 | 8,538 | 167,154 | $34s$ |
| (10, 44, 10) | 14,785 | 1,475 | 17,817 | $4s$ |
| (11, 47, 11) | 16,866 | 6,501 | 153,436 | $50s$ |
| (12, 60, 12) | 27,318 | 14,333 | 111,629 | $33s$ |
| (13, 64, 13) | 32,087 | 7,220 | 103,560 | $2m34s$ |
| (13, 60, 13) | 39,052 | 11,958 | 1,231,516 | $16m26s$ |
| (13, 65, 13) | 36,040 | 11,366 | 183,664 | $49s$ |
| (15, 70, 15) | 28,080 | 14,546 | 997,728 | $3m26s$ |
| (16, 72, 16) | 31,525 | 25,736 | 1,023,012 | $6m13s$ |
| (18, 84, 18) | 50,306 | 21,582 | 524,416 | $4m5s$ |
| (19, 88, 19) | 130,168 | 7,286 | 140,952 | $46s$ |
| (20, 84, 20) | 25,012 | 16,147 | 232,175 | $1m12s$ |

Table 3: Results for randomly-generated LIMIDs.

### 4.4 Comparison to variable elimination

The state-of-the-art exact algorithm for solving LIMIDs is a recently-developed variable elimination algorithm called Multiple Policy Updating (MPU) [11, 10]. Although it is not possible to draw definite conclusions about relative performance without direct comparison, we can make some general comments. Like our branch-and-bound algorithm, the MPU algorithm avoids solving redundant decision scenarios by caching and reusing intermediate results. Our algorithm also uses bounds to prune decision scenarios before they are evaluated, however, and can prune zero-probability branches that represent impossible scenarios, and that may give it some advantage, similar to the advantage that the depth-first branch-and-bound algorithm for solving traditional IDs has over other ID algorithms [24]. In reporting results for their variable elimination algorithm, Maua et al. [11, 10] report that it solves randomly-generated LIMIDs with up to $10^{64}$ strategies and a treewidth bounded by 10. Our branch-and-bound algorithm solves randomly-generated LIMIDs with up to $10^{152}$ strategies and a treewidth of up to 27. (The multi-agent tiger LIMID has $10^{88}$ possible strategies and a treewidth of 38. The 10-stage maze problem has $10^{156}$ possible strategies. The 7-stage maze LIMID has a treewidth of 31; we could not compute the treewidth for more stages than that.) Whereas the scalability of the MPU algorithm is limited by the treewidth of the LIMID, the scalability of the branch-and-bound algorithm appears to be limited by the (usually smaller) treewidth of the relaxed LIMID, which is used to compute bounds and probabilities. In future work, we hope to better characterize the relative performance of these two approaches.

## 5 Conclusion

We have described a branch-and-bound AND/OR graph search algorithm that finds optimal strategies for LIMIDs, building on earlier work on solving traditional IDs using branch-and-bound search. The approach is especially effective for IDs that represent multistage decision problems.

The branch-and-bound approach performs well even though the bounds used in our implementation are quite simple. (They are equivalent to assuming perfect information.) The bounds can likely be significantly improved, allowing more pruning and faster search. We plan to implement improved bounds and evaluate the approach on a wider range of test problems. Our approach to determining the context of a decision node and merging duplicate OR nodes is also simple, and could potentially be improved, and it may also be possible to find more compact representations of an optimal strategy.

**Acknowledgments** This research is partially supported by NSF grants IIS-0953723 and IIS-1219114.